\documentclass[conference]{IEEEtran}

\makeatletter
\def\ps@IEEEtitlepagestyle{%
  \def\@oddfoot{\mycopyrightnotice}%
  \def\@oddhead{\hbox{}\@IEEEheaderstyle\leftmark\hfil\thepage}\relax
  \def\@evenhead{\@IEEEheaderstyle\thepage\hfil\leftmark\hbox{}}\relax
  \def\@evenfoot{}%
}
\def\mycopyrightnotice{%
  \begin{minipage}{\textwidth}
  \scriptsize
  \copyright~2021 IEEE. Personal use of this material is permitted. Permission from IEEE must be obtained for all other uses, in any current or future media, including reprinting/republishing this material for advertising or promotional purposes, creating new collective works, for resale or redistribution to servers or lists, or reuse of any copyrighted component of this work in other works. 
  
  This work has been accepted at The International Conference on Field-Programmable Technology (FPT’21).
  \end{minipage}
}
\makeatother

\usepackage[backend=biber,style=ieee,natbib=true]{biblatex} 
\usepackage{amsmath,amssymb,amsfonts}
\usepackage{graphicx}
\usepackage{textcomp}
\usepackage{xcolor}
\usepackage[para]{footmisc}

\usepackage{booktabs}
\usepackage[binary-units=true]{siunitx}
\usepackage{array,multirow}
\usepackage{hyperref}
\usepackage{amsmath}
\usepackage{enumitem}
\usepackage{algorithm}
\usepackage{subfigure}
\usepackage{graphicx}
\usepackage{tablefootnote}

\usepackage[noend]{algpseudocode}

\newcommand{\figref}[1]{Figure~\ref{#1}}
\newcommand{\secref}[1]{Section~\ref{#1}}

\newcommand{\tabref}[1]{Table~\ref{#1}}

\newcolumntype{L}[1]{>{\raggedright\let\newline\\\arraybackslash\hspace{0pt}}m{#1}}
\newcolumntype{C}[1]{>{\centering\let\newline\\\arraybackslash\hspace{0pt}}m{#1}}
\newcolumntype{R}[1]{>{\raggedleft\let\newline\\\arraybackslash\hspace{0pt}}m{#1}}

\ifCLASSINFOpdf
\else
\fi

\begin{document}
\title{Optimizing Bayesian Recurrent Neural Networks \\ on an FPGA-based Accelerator}

\author{\IEEEauthorblockN{Martin Ferianc\IEEEauthorrefmark{3}\IEEEauthorrefmark{1},
Zhiqiang Que\IEEEauthorrefmark{3}\IEEEauthorrefmark{2},
Hongxiang Fan\IEEEauthorrefmark{4}\IEEEauthorrefmark{2},
Wayne Luk\IEEEauthorrefmark{2}, 
and
Miguel Rodrigues\IEEEauthorrefmark{1}}
\IEEEauthorblockA{\IEEEauthorrefmark{1}Department of Electronic and Electrical Engineering, University College London, London UK,\\ \textit{\{martin.ferianc.19, m.rodrigues\}@ucl.ac.uk}}
\IEEEauthorblockA{\IEEEauthorrefmark{2}Department of Computing, Imperial College London, London UK, \textit{\{z.que, h.fan17, w.luk\}@imperial.ac.uk}}}

\maketitle

\begingroup\renewcommand\thefootnote{\IEEEauthorrefmark{3}}
\footnotetext{Equal contribution.}
\endgroup
\begingroup\renewcommand\thefootnote{\IEEEauthorrefmark{4}}
\footnotetext{Corresponding author.}

\begin{abstract}
Neural networks have demonstrated their outstanding performance in a wide range of tasks. Specifically recurrent architectures based on long-short term memory (LSTM) cells have manifested excellent capability to model time dependencies in real-world data. However, standard recurrent architectures cannot estimate their uncertainty which is essential for safety-critical applications such as in medicine. 
In contrast, Bayesian recurrent neural networks (RNNs) are able to provide uncertainty estimation with improved accuracy. Nonetheless, Bayesian RNNs are computationally and memory demanding, which limits their practicality despite their advantages. To address this issue, we propose an FPGA-based hardware design to accelerate Bayesian LSTM-based RNNs. To further improve the overall algorithmic-hardware performance, a co-design framework is proposed to explore the most fitting algorithmic-hardware configurations for Bayesian RNNs.
We conduct extensive experiments on healthcare applications to demonstrate the improvement of our design and the effectiveness of our framework. Compared with GPU implementation, our FPGA-based design can achieve up to 10 times speedup with nearly 106 times higher energy efficiency. To the best of our knowledge, this is the first work targeting acceleration of Bayesian RNNs on FPGAs.
\end{abstract}

\IEEEpeerreviewmaketitle

\section{Introduction}\label{sec:introduction}

Recurrent neural networks (RNNs) have demonstrated their successes in various sequencing modelling tasks~\cite{han2017ese}.
Among RNN variants~\cite{chang2015recurrent,hochreiter1997long}, Long Short-Term Memory (LSTM) has become the most wide-spread cell due to its ability in utilizing and remembering the past knowledge~\cite{hochreiter1997long}.
Although the regular LSTM-based RNNs show excellent capability in time-series modelling, they are not able to express their model-epistemic uncertainty and they may overfit on small datapools~\cite{gal2016theoretically}.

To enable uncertainty estimation, overfitting prevention and overall accuracy improvement,
Bayesian LSTM-based RNNs have been proposed~\cite{gal2016theoretically}, which learn distributions over their weights instead of constant-pointwise values. Through repeated Monte Carlo (MC) sampling of the weights and corresponding multiple feedforward passes through the network, the Bayesian model is able to express its prediction along with both epistemic and aleatoric uncertainty~\cite{gal2016theoretically}. Bayesian RNNs were applied in contrasting applications, for example: unemployment forecasting~\cite{mcdermott2019bayesian}, fault detection~\cite{sun2020fault}, language modelling~\cite{fortunato2017bayesian} or medicine~\cite{van2017bayesian}.

The deployment of Bayesian RNNs is especially useful in medical applications where uncertainty estimation enables users to better understand and interpret the model's predictions. A demonstration of this is shown in~\figref{fig:introduction_comparison} where the Bayesian recurrent architecture is used to detect anomalies in an electrocardiogram (ECG) through its reconstruction. 
In an anomalous ECG on the right, the model is more uncertain in its prediction in comparison to the normal case.
Therefore, a physician can be better guided in their investigations and diagnoses based on the modelled uncertainty, instead of looking only at the reconstructed mean or the quantitative metrics.
\begin{figure}[t]
 \centering
 \includegraphics[width=1.0\linewidth]{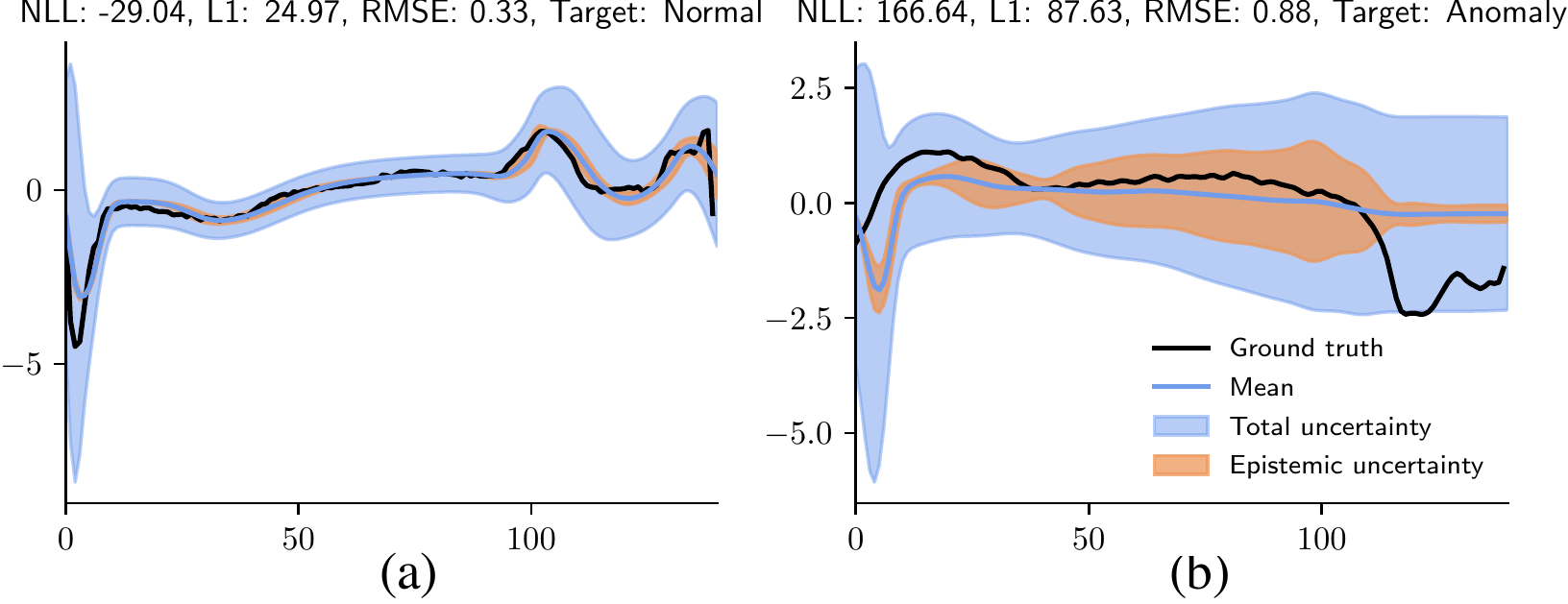}
 \caption{Anomaly detection in a normal (a) and anomalous ECG case (b). The Bayesian model can perfectly fit the normal case, while not being able to replicate the anomalous case along with high uncertainty. {The y-axis represents zero mean and unit variance centered voltage. The x-axis represents the timesteps with 140 timesteps in total.} The fit is measured with respect to negative log-likelihood (NLL), L1 and root-mean-squared error (RMSE). Total uncertainty combines aleatoric and epistemic uncertainty. The predicted uncertainty, as a shaded area, is shown as $\pm$ 3 standard deviations.}
 \label{fig:introduction_comparison}
\end{figure}

However, the benefits of Bayesian RNNs come with real-world execution burdens: the required MC sampling to obtain the prediction as well as the model uncertainty degrade their hardware performance, which limits their deployment in real-life applications. 
For instance, a typical three-layer Bayesian RNN with hidden size being $32$ with $100$ MC samples requires $10.46$ seconds on an Intel Xeon CPU, which cannot meet the requirements of real-world applications, e.g. with respect to real-time ECG analysis~\cite{van2017bayesian} or fault detection~\cite{sun2020fault}. 

Therefore, there is a demand for specific hardware accelerators for Bayesian RNNs. Nevertheless, there are several challenges while accelerating Bayesian RNNs:
\begin{itemize}[leftmargin=*]
    \item \textit{Compute-intensive:} To make a prediction, Bayesian RNN might sequentially perform the feedforward pass through the whole network $S$ times, which significantly increases the amount of required computation.
    \item \textit{Memory-intensive:} Sampling the weight distributions $S$ times produces $S$ different sets of weights, which multiplies the memory requirement by $S$ times compared with that of pointwise non-Bayesian RNNs.
    \item \textit{Resource-intensive:} As Bayesian RNN requires to implement both an RNN engine and random number generators, it demands more resources than a pointwise alternative.
\end{itemize}

{ In this work we introduce several strategies to target these challenges. The compute and memory demands are targeted by our proposed pipelining scheme and efficient random number generation that account for recurrence and data dependency of Bayesian RNNs. Moreover, the structure and portion of Bayesian layers in an RNN and the configuration of our hardware design present a trade-off between algorithmic and hardware performance. To provide efficient resource utilization, we introduce a framework for design space exploration (DSE) tailored to Bayesian RNNs and a configurable accelerator. To the best of our knowledge, this is the first field-programmable gate array (FPGA) based accelerator for Bayesian LSTM-based RNN architectures using Monte Carlo Dropout (MCD)~\cite{gal2016theoretically}.}
In summary, our contributions include:

\begin{itemize}[leftmargin=*]
  \item A novel hardware architecture to accelerate Bayesian LSTM-based recurrent neural networks inferred through Monte Carlo Dropout, which achieves low latency and high energy efficiency~(\secref{sec:hardware}). 
  \item An automatic framework for exploring the algorithmic-hardware performance trade-off under users' requirements e.g. with respect to uncertainty estimation while targeting Bayesian recurrent architectures~(\secref{sec:framework}). 
  
  \item A comprehensive evaluation of algorithmic and hardware performance with respect to real-time ECG anomaly detection and classification with respect to different LSTM-based recurrent architectures~(\secref{sec:experiments}).
\end{itemize}

\section{Preliminaries and Related Work}\label{sec:related_work}

In this section we review recurrent neural networks, Bayesian inference and related hardware accelerators.

\subsection{Recurrent Neural Networks}\label{seq:related_work_lstm}
RNNs were demonstrated to achieve outstanding performance in a number of tasks where understanding time-related relationships was crucial~\cite{hochreiter1997long, srivastava2015unsupervised, hou2019lstm}. In particular, LSTM~\cite{hochreiter1997long} was proven to be effective in capturing long-term dependencies through recurrent processing and storing of useful information. Therefore, this work focuses on accelerating recurrent architectures built around LSTMs. 
LSTM operation can be described by the following equations:
\begin{small}
\begin{align*}
\boldsymbol{i}_t &= \sigma(\boldsymbol{W}^i_x\boldsymbol{x}^i_{t} +\boldsymbol{W}^i_h\boldsymbol{h}^i_{t-1} + \boldsymbol{b}^i)          & \boldsymbol{c}_t &= \boldsymbol{f}_t \odot \boldsymbol{c}_{t-1} + \boldsymbol{i}_t \odot \boldsymbol{g}_t \\
\boldsymbol{f}_t &= \sigma(\boldsymbol{W}^f_x\boldsymbol{x}^f_{t} +\boldsymbol{W}^f_h\boldsymbol{h}^f_{t-1} + \boldsymbol{b}^f)         & \boldsymbol{h}_t &= \boldsymbol{o}_t \odot tanh(\boldsymbol{c}_t)\\
\boldsymbol{g}_t &= tanh(\boldsymbol{W}^g_x\boldsymbol{x}^g_{t} +\boldsymbol{W}^g_h\boldsymbol{h}^g_{t-1} + \boldsymbol{b}^g)    \\
\boldsymbol{o}_t &= \sigma(\boldsymbol{W}^o_x\boldsymbol{x}^o_{t} +\boldsymbol{W}^o_h\boldsymbol{h}^o_{t-1} + \boldsymbol{b}^o)
\end{align*}
\end{small}
The $\sigma,\ tanh,\ \odot$ represent element-wise sigmoid, tanh and multiplication operations. $\boldsymbol{W}=\{\boldsymbol{W}_x^i, \boldsymbol{W}_x^f, \boldsymbol{W}_x^g, \boldsymbol{W}_x^o,$ $\boldsymbol{W}_h^i, \boldsymbol{W}_h^f, \boldsymbol{W}_h^g, \boldsymbol{W}_h^o\}$ and $\boldsymbol{b}=\{\boldsymbol{b}^i, \boldsymbol{b}^f, \boldsymbol{b}^g, \boldsymbol{b}^o\}$ represent the learnable weights and biases. $\boldsymbol{x}_{t}, \boldsymbol{h}_{t-1}, \boldsymbol{c}_{t-1}$ denote the input $\boldsymbol{x}_{t} \in \mathbb{R}^I$ with $I$ features, hidden state $\boldsymbol{h}_{t-1} \in \mathbb{R}^H$ with $H$ features and the cell state $\boldsymbol{c}_{t-1} \in\mathbb{R}^H$ with $H$ features at the current time step $t$ or $t-1$, the previous time step out of total time steps $T$. Note that, $\boldsymbol{h}_0, \boldsymbol{c}_0$ are initialized as zeroes. The intermediate outputs $\boldsymbol{i}_t, \boldsymbol{f}_t, \boldsymbol{g}_t, \boldsymbol{o}_t\in \mathbb{R}^{H}$ are the input, forget, modulation and output gates respectively. Note that we replicate the input $\boldsymbol{x}_{t}$ and the hidden state $\boldsymbol{h}_{t-1}$ such that:
\begin{align*}
    \boldsymbol{x}^i_{t}, \boldsymbol{x}^f_{t}, \boldsymbol{x}^g_{t}, \boldsymbol{x}^o_{t} &= \boldsymbol{x}_{t}\\ 
    \boldsymbol{h}^i_{t-1}, \boldsymbol{h}^f_{t-1}, \boldsymbol{h}^g_{t-1}, \boldsymbol{h}^o_{t-1} &= \boldsymbol{h}_{t-1}
\end{align*}
The decoupling of the input and the hidden state for each weight or gate is crucial for performing Bayesian inference~\citep{gal2016theoretically}.

\subsection{Bayesian Inference}\label{sec:related_work_bayesian_inference}
MCD in RNNs lays in casting dropout~\citep{srivastava2014dropout} as Bayesian inference with two major differences~\citep{gal2016theoretically}. First, the dropout is enabled during training as well as evaluation. Second, the dropout mask $\boldsymbol{z}\sim \textrm{Bernoulli}(1-p); \boldsymbol{z}=\{\boldsymbol{z}^i_x, \boldsymbol{z}^f_x, \boldsymbol{z}^g_x, \boldsymbol{z}^o_x \in \mathbb{R}^{I}; \boldsymbol{z}^i_h, \boldsymbol{z}^f_h, \boldsymbol{z}^g_h, \boldsymbol{z}^o_h\in \mathbb{R}^{H}\}$, with the same dimensionality as one time step of the input or the hidden state, is sampled only once for all time steps $T$ and individually for all $\boldsymbol{x}^i_{t}, \boldsymbol{x}^f_{t}, \boldsymbol{x}^g_{t}, \boldsymbol{x}^o_{t}$ and  $\boldsymbol{h}^i_{t-1}, \boldsymbol{h}^f_{t-1}, \boldsymbol{h}^g_{t-1}, \boldsymbol{h}^o_{t-1}$, such that for example $\boldsymbol{x}^i_{t} = \boldsymbol{x}^i_{t} \odot \boldsymbol{z}^i_x$ or $\boldsymbol{h}^i_{t-1} = \boldsymbol{h}^i_{t-1} \odot \boldsymbol{z}^i_h$. 
Probability $p\in [0, 1]$ of sampling 0 practically represents the trade-off between accuracy and calibration of the architecture. Dropout can be applied to only input, only the hidden states or both and it does not need to be applied to every cell in an architecture, which results in a partially Bayesian architecture~\citep{kendall2015bayesian}. From the hardware perspective, such architectures represent a trade-off between algorithmic and hardware performance~\citep{fan2021highperformance}. The prediction in Bayesian architectures is obtained by running the same input through the RNN $S$ times, each time with a different set of sampled masks $\boldsymbol{z}$ for each layer $i$ where MCD is applied. The collected outputs from the individual passes are then averaged to form a prediction. The $S$ samples increase the compute and number of memory accesses linearly with complexity $\mathcal{O}(S)$. Bayesian RNNs have been used in time-series forecasting and classification~\citep{mcdermott2019bayesian,van2017bayesian}, where they demonstrated fine algorithmic performance.

\subsection{Hardware Accelerators}
Due to high computational, low-latency and reconfigurability demands, custom hardware accelerators for NNs represent a viable implementation platform. Especially FPGAs present an energy-efficient, configurable and high-performance hardware technology for accelerating NN architectures~\citep{mittal2020survey}. 

There has been ample work on FPGA-based implementations of persistent LSTMs whose weights are stored in on-chip memory~\cite{ rybalkin2018finn, eriko_fccm19,que2020reconfigurable, rybalkin2020efficient, rybalkin2020massive}. 
For example, \textit{FINN-L}~\cite{rybalkin2018finn} quantizes the RNN into 1-8 bits which surpasses a single-precision floating-point accuracy for a given dataset. 
Other studies on LSTM implementations store weights in the off-chip memory considering an FPGA, which was identified as a performance bottleneck~\cite{guan2017fpga, chang2015recurrent, que2020mapping,park2020time}. In addition, LSTM weights' reuse methods~\cite{que2020mapping, park2020time} between various timestep were proposed to reduce the off-chip memory accesses to decrease the energy cost and improve the overall system's throughput. Some of the previous studies~\cite{han2017ese, cao2019efficient, shi2019lstm, nan2020dc} focused on weight pruning and model compression to reduce the size of weights to achieve favorable hardware performance. In~\cite{chen2020blink}, \textit{BLINK} was proposed which utilized bit-sparse data representation for the LSTM runtime. It improved the energy efficiency of the LSTM by turning the multiplication into a bit shift operation without impairing its accuracy. However, none of these FPGA-based RNN designs target Bayesian RNNs.

At the same time, several hardware accelerators have been proposed to accelerate Bayesian NNs (BNNs)~\citep{jia2020efficient, cai2018vibnn, fan2021highperformance, awano2020bynqnet}.
However, these designs only focus on accelerating feedforward BNNs.
Cai \textit{et al.}~\citep{cai2018vibnn} proposed a hardware design called \textit{VIBNN} to accelerate BNNs consisting only of dense layers. Their accelerator consumes a large amount of resources while implementing Gaussian random number generators.
Awano \& Hashimoto~\citep{awano2020bynqnet} proposed \textit{BYNQNet} to accelerate BNNs, which achieves $4.07$ and $8.99$ times higher throughput and energy efficiency than \textit{VIBNN}.
However, the design puts strict restrictions on the used nonlinear activation functions and thus limiting real-world applicability.
By exploiting the activation sparsity in BNNs,~\citep{wan2020fast} proposed a novel hardware architecture called \textit{Fast-BCNN}.
Nevertheless, the design can only be used to accelerate Bayesian  convolutional NNs (BCNNs) with ReLU~\citep{agarap2018deep}.
Fan \textit{et. al}~\citep{fan2021highperformance} proposed an FPGA-based accelerator for BCNNs inferred through MCD~\citep{gal2016theoretically}.
The design achieves nearly $10$ times higher compute efficiency than \textit{BYNQNet}. 
None of the previously mentioned accelerators for BNNs target RNNs, and thus, they do not consider the recurrence or inherent data dependency in RNNs. 

In comparison to previous work, this paper focuses on accelerating Bayesian RNNs. To the best of our knowledge, this is the first work to accelerate Bayesian RNNs on an FPGA.

\section{Hardware Design}\label{sec:hardware}

In this section we outline the proposed pipelined accelerator and an efficient random number generation for MCD-based BNNs and the target recurrent architectures. 
\begin{figure}
\centering
\includegraphics[width=.9\linewidth]{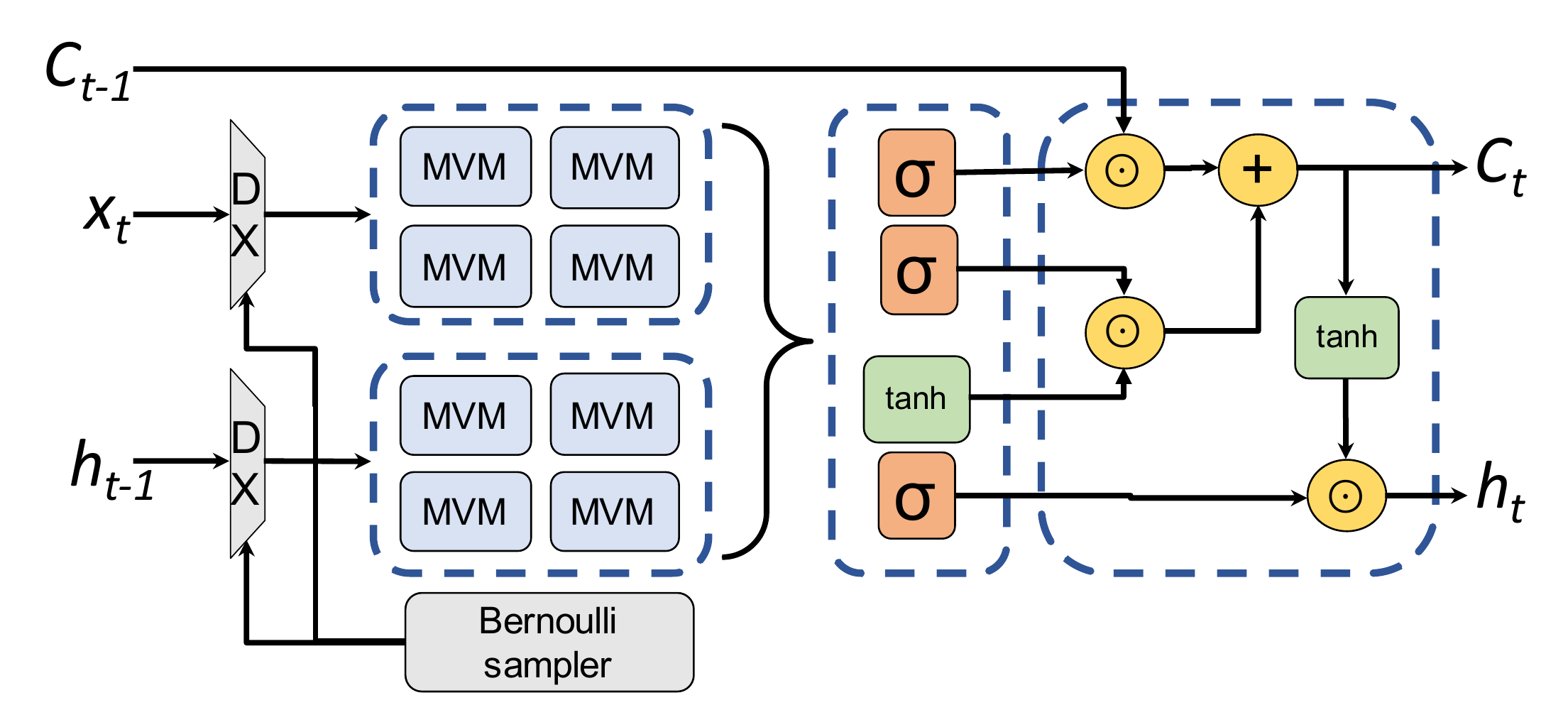}

\caption{Overview of the hardware implementation of the Bayesian LSTM.}
\label{fig:hardware_overview}
\end{figure}

\subsection{Design Overview}
This work adopts a streaming design~\cite{guan2017fpga, duarte2018fast, tridgell2019unrolling} where all individual layers are mapped on-chip and different layers run in a pipelined fashion to achieve low latency.
Unrolling the overall architecture in this way results in a more efficient utilization of resources, with a 1-to-1 ratio of DSP blocks to compute units. Besides, this design adopts the initiation interval (II) balancing for multiple LSTM layers to achieve low latency and high hardware efficiency.

An overview of the proposed hardware design of a single LSTM layer 
is illustrated in~\figref{fig:hardware_overview}. The input and output data are transferred using DMA via an AXI bus.
The input $\boldsymbol{x}_t$ and hidden state $\boldsymbol{h}_{t-1}$ are masked by the output of Bernoulli samplers and then fed to the LSTM gates. The masking along with the decomposition, as discussed in~\secref{seq:related_work_lstm}, is performed by demultiplexor units (DX) that control which individual features get passed forward. 
There are four gates at the front of the LSTM layer, each containing a matrix-vector multiplication (MVM) unit. The element-wise operations and activation functions: sigmoid, tanh, addition and multiplication are performed on the output of the MVMs, previously factoring in the weights $\boldsymbol{W}_i$ and biases $\boldsymbol{b}_i$ of that given LSTM $i$. At the end of the layer, the current cell state $\boldsymbol{c}_t$ and hidden state $\boldsymbol{h}_t$ are produced. 
The $\boldsymbol{h}_t$ is required in the LSTM gates in the next time step iteration; it shows the existence of data dependencies which are not in forward-only NNs.
The activation functions are implemented using BRAM-based lookup tables with a range of precomputed input values.
The weights and biases are mapped on-chip automatically into registers when the design is synthesized. Hence, weight sampling is avoided along with additional memory traffic by introducing routing through DXs, enabling complete on-chip computation and elimination of the memory challenge in~\secref{sec:introduction}. A similar design logic as presented here can be used for other recurrent units such as the gated recurrent unit~\cite{cho2014learning}. 

\subsection{Bernoulli Sampler and Design Pipelining}\label{sec:hardware_dropout}
\begin{figure}
\centering
\includegraphics[width=.5\linewidth]{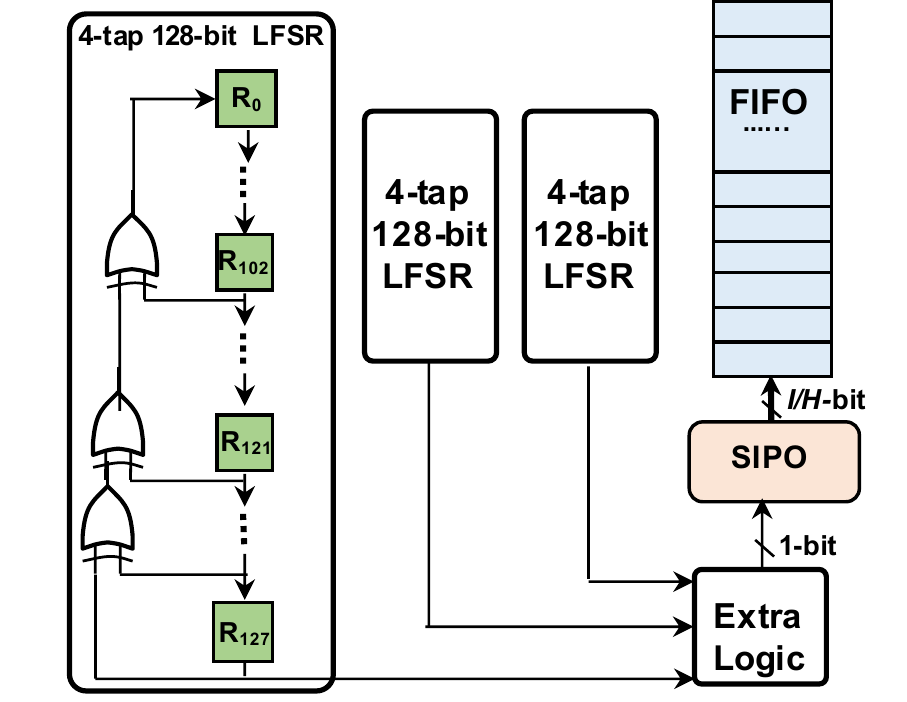}

\caption{Hardware architecture of the implemented Bernoulli sampler.}
\label{fig:hardware_bernoulli}
\end{figure}
As MCD randomly sets inputs as zeros during runtime, it requires the hardware to generate random $1$s and $0$s.
To achieve this goal, we design a Bernoulli sampler in hardware as illustrated in~\figref{fig:hardware_bernoulli}.
The 4-tap linear feedback shift register (LFSR) is the basic module in our Bernoulli sampler, which generates random binary values with a probability of $p$=0.5. 
To generate random binaries with user-defined probability, there are $N_{lfsr}$ LFSRs followed by an extra logic block.
For instance, to generate zeros with a probability $p$=0.125, it requires $N_{lfsr}$=3 with an extra three-input NAND gate as the extra logic.
In this paper, to save the hardware resources, we set $N_{lfsr}$=3 and we set the dropout probability uniformly to $p$=0.125, as advised by~\cite{van2017bayesian}, for both the inputs $\boldsymbol{x}$ as well as hidden states $\boldsymbol{h}$.
Since all the generated random binary values need to be outputted in parallel, a serial-in-parallel-out (SIPO) module is placed after LFSRs followed by a first-in-first-out (FIFO) module. If the given layer is not Bayesian, both DX and Bernoulli sampler are not needed.

\begin{figure}
\centering
\includegraphics[width=1.0\linewidth]{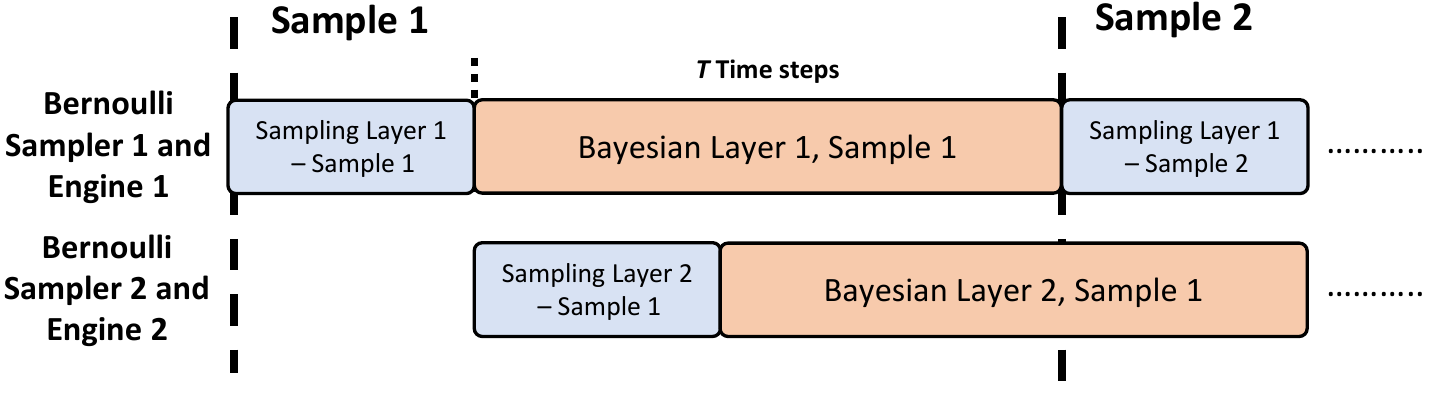}
\caption{Overlapping the computation with Bernoulli sampling.}
\label{fig:overlap_bernoulli}
\end{figure}
To further improve the hardware performance, we propose to overlap the Bernoulli sampling with the computation of LSTMs, which is illustrated in~\figref{fig:overlap_bernoulli}.
As the Bernoulli sampling does not rely on the inputs, it can be performed before the start of all time steps $T$ for a single LSTM.
However, generating random binaries for all engines and inputs and hidden states will cost a large amount of on-chip memory.
Therefore, all the Bernoulli samplers in our design only pre-sample random binaries required by a single input.
This overlapping approach can hide the time cost of Bernoulli sampling into the computation, and at the same time, decrease the on-chip memory consumption. In addition to the sample-wise pipelining, we also introduce the pipelining over each time step to further increase parallelism, as shown in~\figref{fig:ts_pipeline}. Note that while the illustrated example shows 3 cascaded LSTM layers with 4 time steps, the real design may involve more layers and more time steps.
By leveraging the combination of sample-wise pipelining and time step pipelining, our design provides a fundamental solution to the Bayesian RNNs that demand repeated MC sampling, targeting the compute-intensive challenge mentioned in~\secref{sec:introduction}.
\begin{figure}[t]
    \centering
    \includegraphics[width=0.5\linewidth]{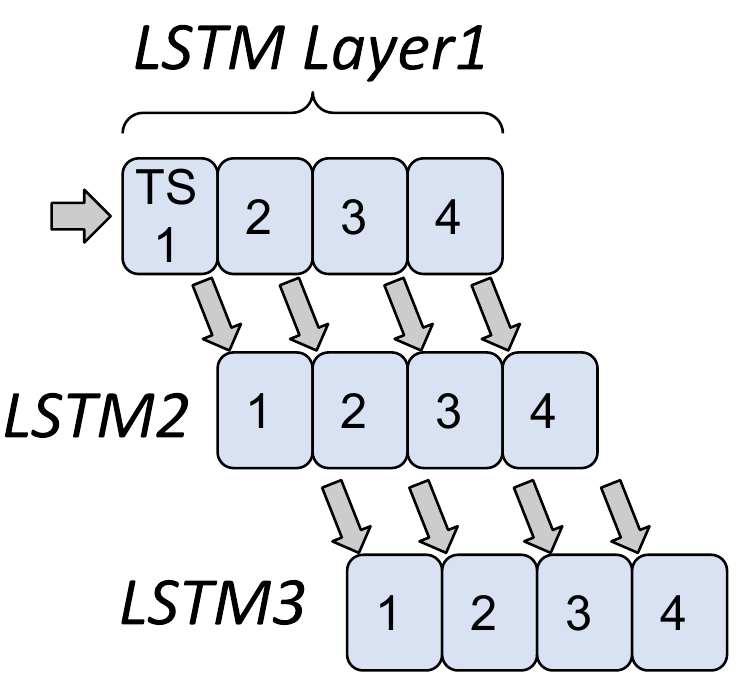}
    \caption{Three cascaded LSTM layers with time step (TS) pipelining.}
    \label{fig:ts_pipeline}
\end{figure}
\begin{figure}[t]
    \centering
    \includegraphics[width=1.0\linewidth]{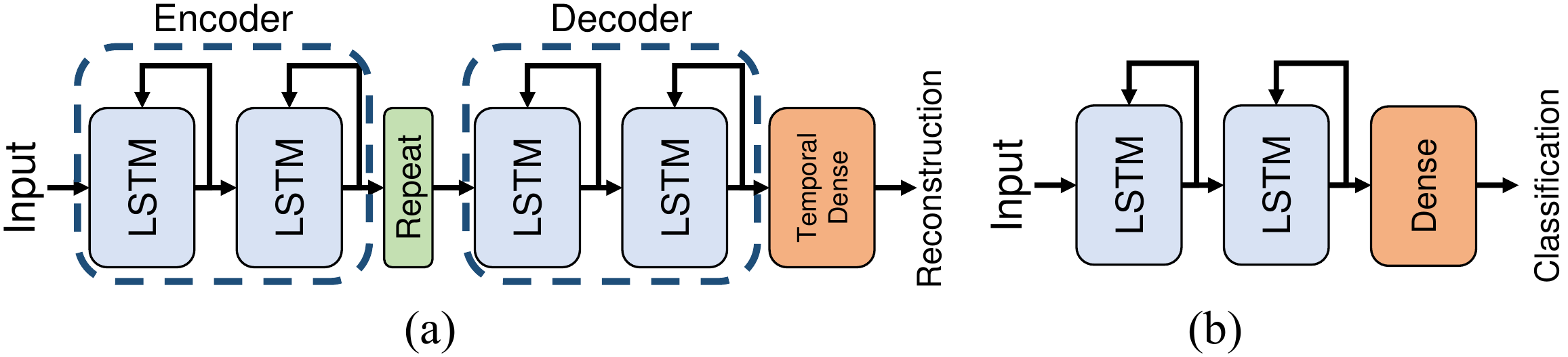}
    \caption{Recurrent autoencoder (a) and classifier (b) architectures each with $NL$=2 LSTMs in their respective parts.}
    \label{fig:algorithmic_architecture}
\end{figure}

\subsection{Recurrent Autoencoder and Classifier}\label{sec:hardware_algorithmic_architectures}
\figref{fig:algorithmic_architecture} (a) demonstrates the hardware architecture of the recurrent autoencoder that is used for anomaly detection~\cite{srivastava2015unsupervised,hou2019lstm} in our experiments. It consists of two parts: a pipelined encoder and a pipelined decoder, each balanced with $NL$ LSTM instances giving in total $2NL$ layers. Given the unrolling, the hardware resource consumption scales with the total layer count. Encoder processes the time-series input $\boldsymbol{x}\in\mathbb{R}^{T\times I}$ into a bottleneck encoding containing only the last hidden state $\boldsymbol{h}_T\in \mathbb{R}^{H/2}$ of the last LSTM in the encoder. {The last hidden state in the decoder has a reduced dimensionality $\mathbb{R}^{H/2}$ in order to learn to convey only the most relevant information to the decoder~\cite{kingma2013auto}.} The encoding is repeated $T$ times which can be effectively achieved by caching it for exactly $T$ time steps. The decoder transforms the repeated embedding time step by time step into output $\boldsymbol{h}\in \mathbb{R}^{T\times H}$. $\boldsymbol{h}$ is then processed further by a temporal dense layer, where the same dense layer processes each output in the sequence to give the final reconstruction of the input, again in a pipelined fashion. A dense layer is simply implemented as a single MVM unit. The architecture aims to learn a useful embedding which captures the essence of the input signal and allows its efficient reconstruction. Based on the quality of the reconstruction the input is labelled as normal or anomalous.

The hardware model for classification can be built in a similar fashion, by considering only the encoder part of the architecture as shown in Figure~\ref{fig:algorithmic_architecture} (b) with $NL$ layers. The last hidden state $\boldsymbol{h}_T\in \mathbb{R}^{H}$ of the encoder is not repeated but processed through a dense layer that reshapes it to the number of output classes and processes it through a softmax activation. Hence the classifier is fully pipelined. Given an input signal, the encoder captures variable-size input relationships into a consistent output embedding that is used for classification, given that the input is labelled. 
The hidden size $H$, number of layers $NL$, the portion of Bayesian LSTMs $\boldsymbol{B}$ or the hardware configuration can vary as we discuss in the optimization framework. 
\section{Optimization Framework}\label{sec:framework}
In this section, we first present an overview of the proposed optimization framework.
Then, the resource and latency models are introduced, which are used to accelerate the DSE.

\begin{figure}[t]
\centering
\includegraphics[width=0.85\linewidth]{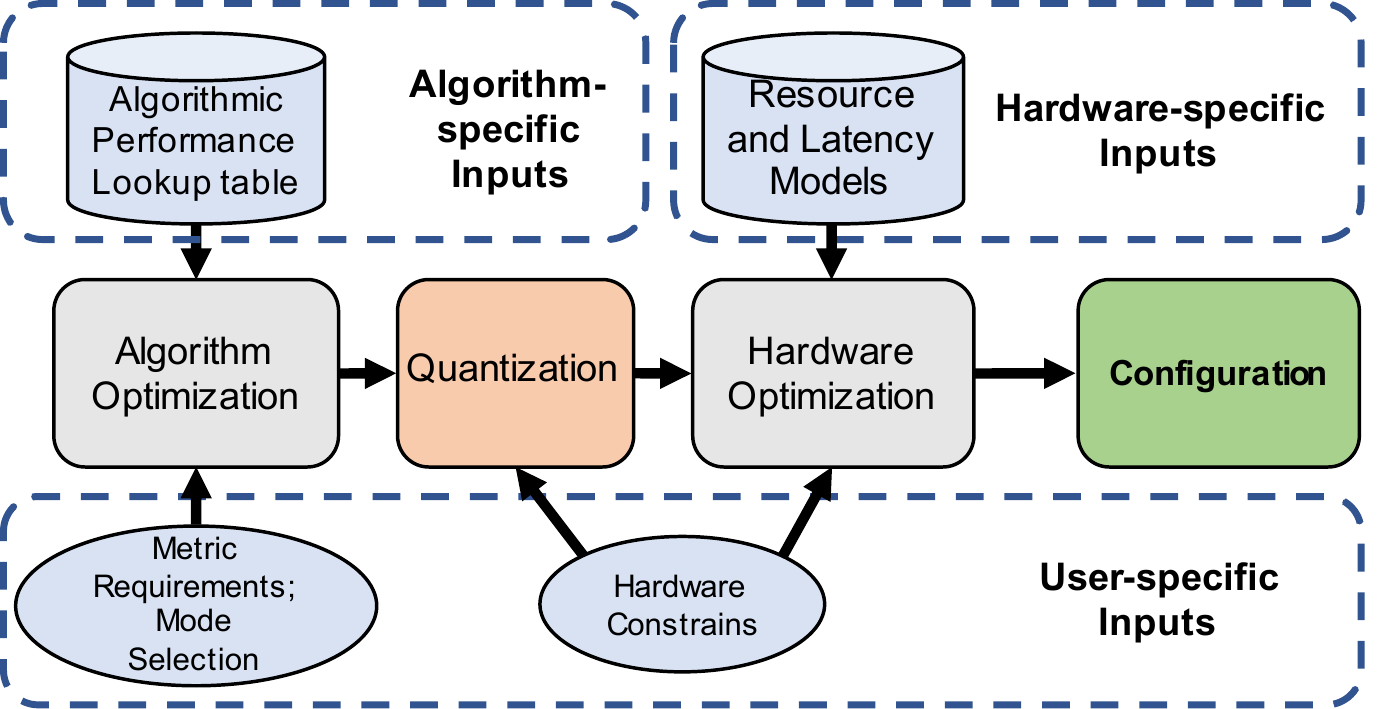}
\caption{Overview of the optimization framework.}
\label{fig:framework_overview}
\end{figure}
\subsection{Overview of Framework}
An overview of the proposed framework is shown in Figure~\ref{fig:framework_overview}.
Given user-defined priorities in terms of the target metric and the platform-specific hardware constraints, it is necessary to optimize both the algorithmic and hardware configuration of the RNN as well as the accelerator.
Therefore, we propose an optimization framework to perform DSE under both user-defined algorithmic and hardware constraints.
In our design, the performance trade-off is decided by two categories of parameters: \textit{1)} Algorithmic architectural parameters, which include the overall network architecture $\boldsymbol{A}$: the hidden size $H$, the number of layers $NL$ for encoder or decoder and the portion of Bayesian layers $\boldsymbol{B}$ and \textit{2)} Hardware parameters $\boldsymbol{R}$: which consist of reuse factors $R_x, R_h, R_d$ of processing engines.
The objective of our framework is to optimize the latency and algorithmic metrics such as accuracy and quality of uncertainty prediction by exploring both $\boldsymbol{A}$ and $\boldsymbol{B}$ to target all three challenges mentioned in~\secref{sec:introduction}.

At the start, the framework requires users to specify the hardware constraints, metric requirements and the focus mode.
The main hardware constraint is the number of available DSPs on the target hardware platform.
The optimization mode is selected to minimize or maximize the chosen objective through greedy optimization with respect to algorithmic and hardware configurations. At first, the algorithmic optimization is conducted with respect to a previously built lookup table consisting of algorithm-benchmarked architectures. Following algorithm optimization and potential re-training, the networks are quantized depending on hardware constraints. In this work we consider 16-bit fixed-point quantization. Next, the parameters $\boldsymbol{R}$ of a hardware configuration are optimized with respect to a hardware model. The hardware model is used to estimate the resource consumption or latency given the available configurations.
Based on the determined hardware parameters, the latency is estimated given a performance lookup table for various BNN configurations with different $\boldsymbol{A}$. At the end, the configurations which do not meet the minimal requirements are filtered resulting in a final configuration.

\subsection{Resource Model}\label{sec:res_model}
In this paper, we mainly consider the resource consumption in terms of DSPs as $DSP_{design}$, which represent the resource bottleneck, while being limited by the total available DSPs as $DSP_{total}$. The number of DSPs for a given LSTM layer $DSP_i$ and the complete design using 16-bit representation, except $\boldsymbol{c}_{t-1}^i$ which is represented in 32-bit, is shown as:
\begin{align}
 DSP_{i} &= \frac{4\times I_{i} \times H_{i}}{R_x} + \frac{4\times H_{i}^2 }{R_h} + 4\times H_{i} \nonumber \\
 DSP_{design} &= \sum_{i = 1}^{L} DSP_{i} + DSP_d\leq DSP_{total} \nonumber
\end{align}
$I_{i}$, $H_{i}$ and $O$ represent the input, hidden state and output dimensionality for layer $i$. $L$ is $2NL$ if considering autoencoder or $NL$ if considering the classifier. The factor of $4$ means there are $4$ MVMs for input and $4$ for the hidden state in a single LSTM layer. The $\boldsymbol{f}_t \times \boldsymbol{c}_{t-1}$ in the LSTM tail needs two Xilinx DSPs to construct one multiplier unit, thus the LSTM tail unit consumes $4\times H_{i}$ DSPs. 
The $DSP_d$ is the DSP consumption for the final dense layer which equals $\frac{H_L \times O \times T}{R_d}$ if considering autoencoder or $\frac{H_L \times O}{R_d}$ if considering the classifier. The $R_x$,  $R_h$ and $R_d$ represent the reuse factors for the MVMs processing the input, hidden state or the final dense layer respectively. $T$ is the time step or sequence length. In the design space exploration, additional 5\% of the $DSP_{total}$ was added since we found that the HLS tool often optimizes the DSP usage by replacing the multipliers using other simpler logic when possible. 

The trade-off between latency, throughput and FPGA resource usage is determined by the parallelism of the calculation.
This work adopts the reuse factor used in~\cite{duarte2018fast} to fine tune the parallelism, which is configured to set the number of times a multiplier is used in the computation of a module. With a reuse factor of $R$, $\frac{1}{R}$ fewer multipliers and computation are performed. With a higher reuse factor, the DSP resource usage can be reduced, however, the latency of processing MVMs will increase. The reuse factors should be carefully chosen so that the design can fit into the targeted FPGA chip while keeping the latency as small as possible.

\subsection{Latency Model}\label{sec:latency_model}

The individual layer latency $Lat_{i}$ and end-to-end latency $Lat_{design}$, which is dominated solely by the recurrent cells, are modeled as:
\begin{align*}
    II &= \max_{i=1,\ldots,L}II_i & Lat_{i} &= II \times T + (IL_{i} - II)
\end{align*}
\vspace{-2em}
\begin{align*}
    Lat_{design} &= II \times T + (IL_{i} - II) \times NL
\end{align*}
where $II$ is the initiation interval of the single time step loop, $T$ is the time sequence length, $IL_{i}$ is the iteration latency. The $II$ is the number of clock cycles before a unit can accept new inputs and is generally the most critical performance metric in many systems~\cite{xilinx_sdsoc}.
In the proposed pipelined design, the processing of the cascaded LSTM layers can be overlapped. For example, the second layer $i$=2 does not need to wait for the whole sequence of hidden states to begin computation. Just a single hidden state from the former LSTM layer is sufficient for the computation in the next LSTM layer. Furthermore, since the $II$ of a model is decided by the largest individual layer $i$, the $II$ of the cascaded layers is set to be the same to achieve the best hardware resource efficiency. Thus, the total latency of a design with $NL$ cascaded LSTM layers is given as $Lat_{design}$. 
It has to be noted that the decoder in the autoencoder can only be started after the encoder calculation is completed, since only the last time step hidden state is returned in the last layer of the encoder. Thus, the latency of an autoencoder with $2NL$ LSTM layers, $NL$ for encoder and $NL$ for decoder, is simply $Lat_{design}\times 2$. 

As shown in the equations above, the $Lat_{design}$ is dominated by the $II$ when $T$ is fixed. 
This work achieves the optimal $II$ via identifying the proper reuse factors under the hardware resource limitations, as shown in~\secref{sec:res_model} to achieve the lowest $II$ and end-to-end latency.

\section{Experiments}\label{sec:experiments}

In this section we first review the general experimental setup followed by algorithmic DSE and hardware performance comparison with respect to different hardware platforms.

The experiments were performed on \textit{ECG5000} dataset~\citep{goldberger2000physiobank}, that contains 5000 samples split into a training set of only 500 samples and a test set of 4500 samples. By default, the dataset has 4 classes: 1 normal and 3 anomalous. Each ECG has $T$=140 and it was preprocessed such that each sample was zero mean and unit variance centered. It is a challenging dataset mainly due to its small size and class imbalance, which can be associated with anomaly detection or classification tasks. For both tasks we trained various recurrent architectures with respect to 1000 epochs, batch size 64, gradient clipping set as $3.0$ and weight decay set as $0.0001$ to provide regularization.

Next, we present the algorithmic DSE that represents an algorithmic optimization and population of the lookup table in the proposed framework. It is followed by a hardware optimization and performance comparison. Arrows in Tables and Figures symbolize desired trends and bold values represent the best score.

\subsection{Algorithmic Optimization}\label{sec:experiments_algorithm_optimization}

\subsubsection{Anomaly Detection}\label{sec:experiments_algorithm_optimization_anomaly}
For anomaly detection, we split the data into normal and anomalous samples. We appended anomalous cases from the train set to the test set and we trained the autoencoder from Section~\ref{sec:hardware_algorithmic_architectures} only with respect to normal data to be able to recreate it. We measured the wellness of the fit with respect to root-mean squared error. Based on the fit, we analyzed the models with respect to receiver operating characteristic and the respective area under the curve (AUC), average precision (AP) and accuracy (ACC) at the cutoff point that maximizes true positive rate against false positive rate in detecting anomalies. We considered autoencoders with $\boldsymbol{A}$: $H$=\{8, 16, 24, 32\} and $NL$=\{1, 2\} LSTMs in encoder and decoder with dropout $\boldsymbol{B}$ benchmarked at every position and combination. 
\begin{figure}
    \centering
    \includegraphics[width=0.9\linewidth]{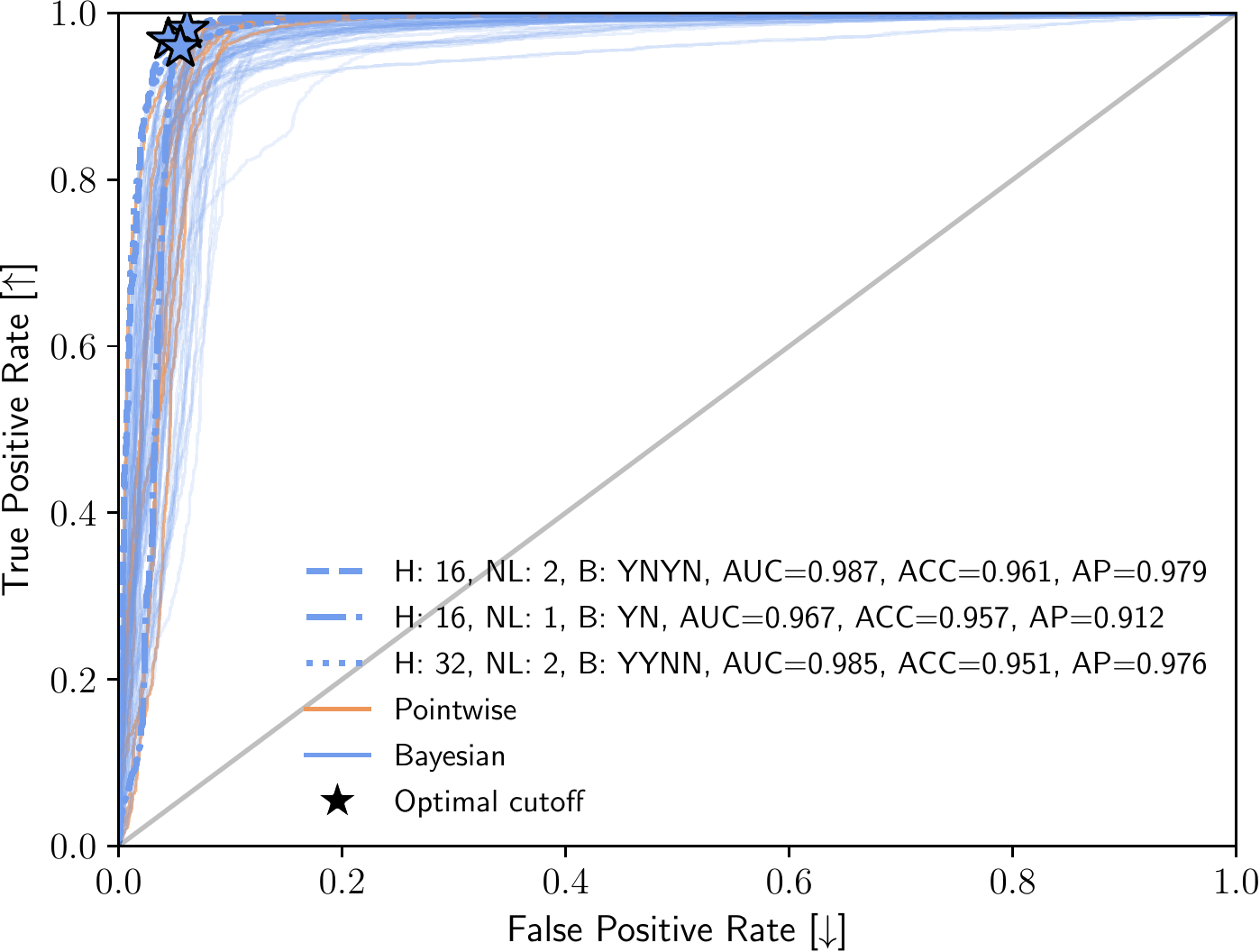}
    \caption{Receiver operating characteristic on the ECG test set with respect to Bayesian and pointwise (without any Bayesian layers) autoencoders in anomaly detection. H is hidden size, NL is number of layers in encoder or decoder, B symbolizes Bayesian in the given layer enabled (Y) or disabled (N). AUC is area under the curve, ACC is accuracy and AP is average precision.}
    \label{fig:dse_anomaly}
\end{figure}
\begin{figure}
    \centering
    \includegraphics[width=0.9\linewidth]{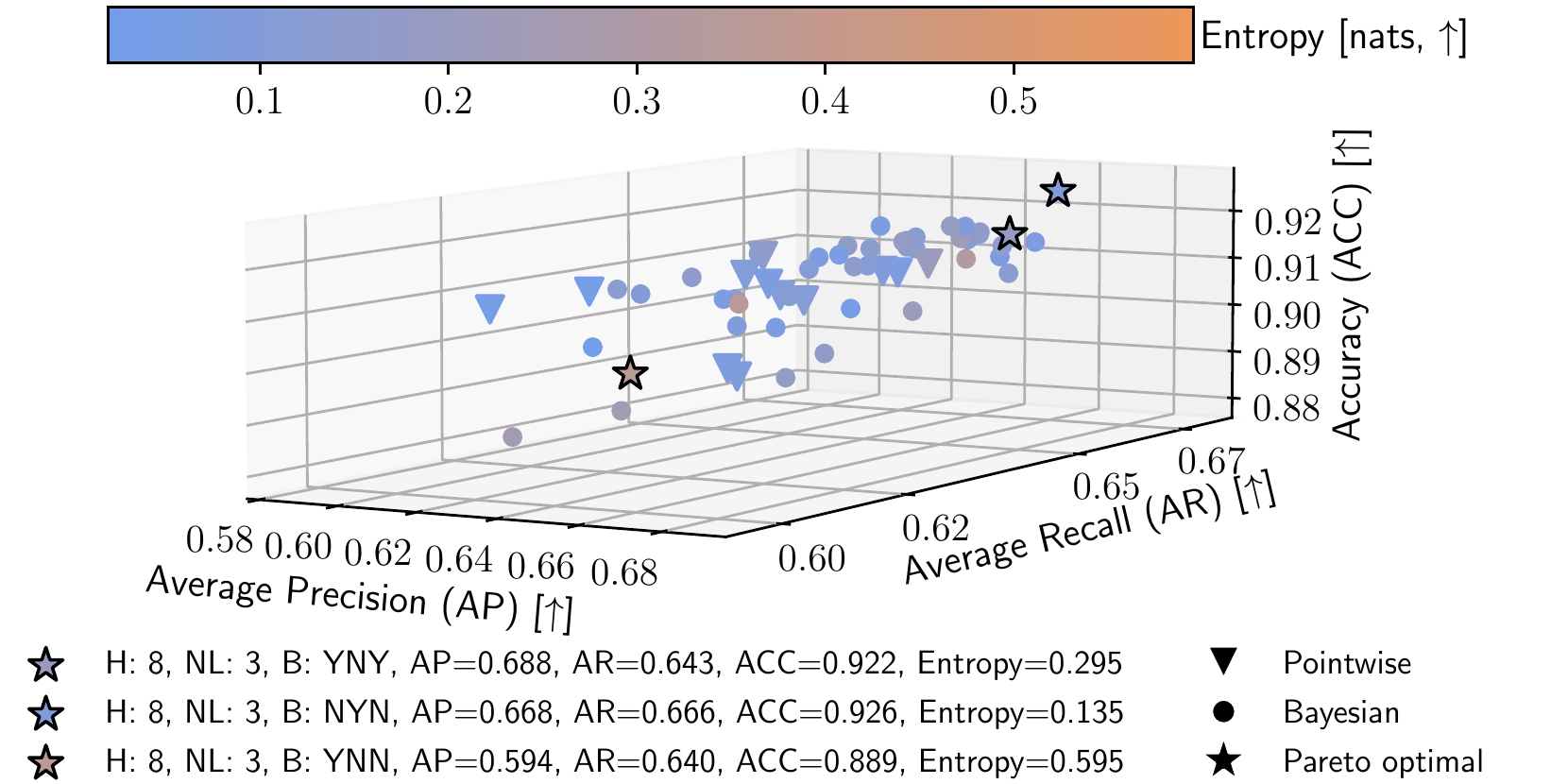}
    \caption{Classification performance on the ECG test set with respect to Bayesian and pointwise (without any Bayesian layers) recurrent nets. H is hidden size, NL is number of layers, B symbolizes Bayesian in the given layer enabled (Y) or disabled (N).}
    \label{fig:dse_classification}
\end{figure}

The results with respect to the DSE in floating-point and $S$=30 are shown in Figure~\ref{fig:dse_anomaly}. It can be seen that the Pareto optimal architectures were at least partially Bayesian. The best model was with $H$=16, $NL$=2 layers in encoder or decoder and dropout applied both in the encoder and decoder $\boldsymbol{B}$=YNYN. Y/N stands for MCD enabled or disabled respectively for that layer. The model achieved fine algorithmic performance with AUC, AP and ACC all approaching 1. 

\subsubsection{Classification}\label{sec:experiments_algorithm_optimization_classification}
For classification we evaluated the models trained on all four classes with respect to ACC, macro AP and average recall (AR), since the dataset is severely unbalanced. Additionally, we considered uncertainty estimation qualities with respect to sequences of random Gaussian noise for which we measured the predictive entropy in nats. We considered classifiers as per Section~\ref{sec:hardware_algorithmic_architectures} with $\boldsymbol{A}$: $H$=$\{8, 16, 32, 64\}$ and $NL$=$\{1, 2, 3\}$ LSTMs in the encoder with dropout $\boldsymbol{B}$ benchmarked at every position and combination. Figure~\ref{fig:dse_classification} summarizes the performance of the considered architectures while running in floating-point and $S$=30. Similarly to anomaly detection, the best performing architectures were again at least partially Bayesian. In this case the optimal architecture was identified with $H$=8, $NL$=3 layers overall with dropout applied such that $\boldsymbol{B}$=YNY. The model achieved high accuracy and precision.

\subsubsection{Sampling} As discussed in Section~\ref{sec:related_work_bayesian_inference}, the software performance of Bayesian architectures depends on the number of feedforward samples $S$ that also affects the overall runtime. Figures~\ref{fig:samples_dse} (a,b) demonstrate the relationship between the software metrics for both anomaly detection (a) and classification (b) with respect to the best architectures for each task. It can be seen that  an $S$ larger than $30$ results in diminishing returns.
\begin{figure}[t]
    \centering
    \includegraphics[width=1.0\linewidth]{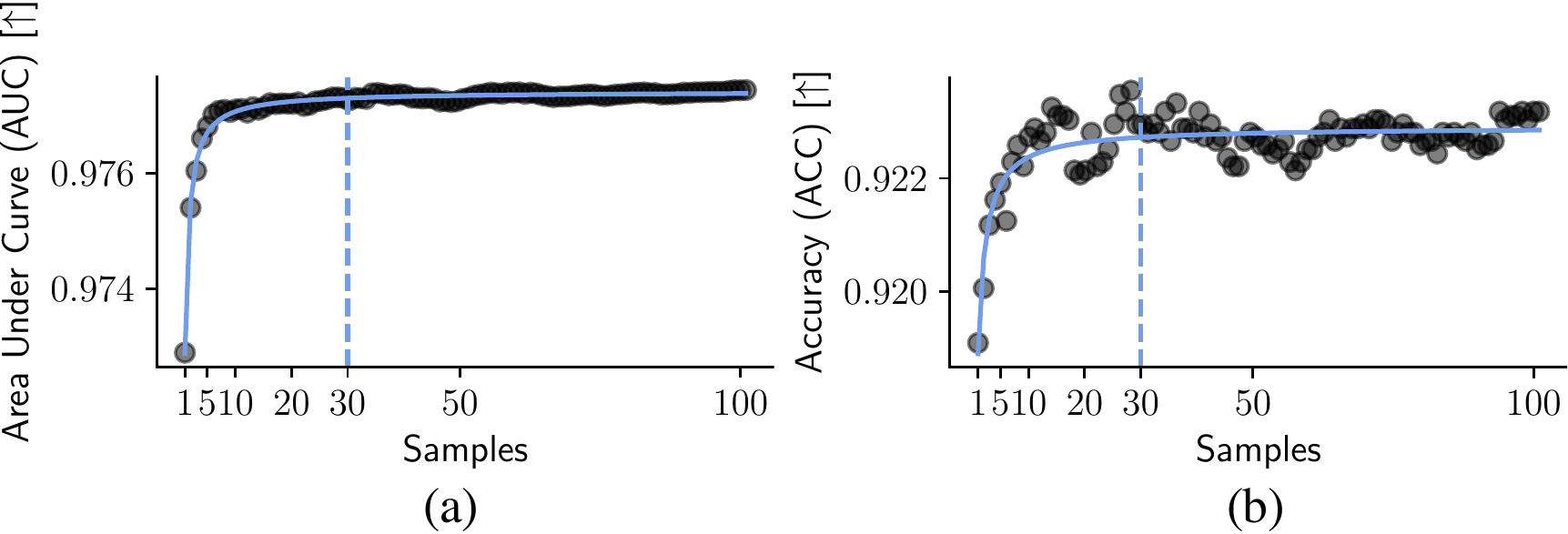}

    \caption{Software performance change for anomaly detection (a) and classification (b) with increasing number of samples $S$ from 1, 30 to 100 samples.}
    \label{fig:samples_dse}
\end{figure}

\subsection{Quantization}\label{sec:experiments_quantization}
The next step given the outlined framework in Section~\ref{sec:framework} is quantization. In Tables~\ref{tab:dse_anomaly_float_quant} and~\ref{tab:dse_classification_float_quant} the performance of the best floating-point models for both anomaly detection and classification is compared with respect to the 16-bit fixed-point quantization when $S$=30. The results were collected with respect to retraining the best architectures three times to obtain the mean and the standard deviation for comparison. The results demonstrate that the chosen fixed-point quantization scheme and configuration preserves high accuracy and uncertainty estimation, seen in entropy, of both best models. 

\begin{table}[t]
  \caption{Comparison of floating-point and quantized best model for anomaly detection.}
  \label{tab:dse_anomaly_float_quant}
  \centering
  \scalebox{1.0}{
  \begin{tabular}{c|c|c|c}
    \toprule
    \begin{tabular}[x]{@{}c@{}}\textbf{Representation}\\ \textbf{Precision}\end{tabular} &
    \textbf{Accuracy} [$\uparrow$]&
    \begin{tabular}[x]{@{}c@{}}\textbf{Average}\\ \textbf{Precision}\end{tabular} [$\uparrow$]&
    \begin{tabular}[x]{@{}c@{}}\textbf{Area} \textbf{under} \\ \textbf{Curve}\end{tabular} [$\uparrow$] \\ 
    
    \midrule
    Floating-point & $0.95 \pm 0.01$ & $0.96 \pm 0.02$ & $0.98 \pm 0.01$ \\
    \midrule
    Fixed-point & $0.95 \pm 0.01$ & $0.97 \pm 0.01$ & $0.98 \pm 0.01$ \\
    \bottomrule
  \end{tabular}}
\end{table}
\begin{table}[t]
  \caption{Comparison of floating-point and quantized best model for classification.}
  \label{tab:dse_classification_float_quant}
  \centering
  \scalebox{.98}{
  \begin{tabular}{c|c|c|c|c}
    \toprule
    \begin{tabular}[x]{@{}c@{}}\textbf{Representation}\\ \textbf{Precision}\end{tabular} &
    \textbf{Accuracy} [$\uparrow$]&
    \begin{tabular}[x]{@{}c@{}}\textbf{Average}\\ \textbf{Precision}\end{tabular} [$\uparrow$]&
    \begin{tabular}[x]{@{}c@{}}\textbf{Average} \\ \textbf{Recall}\end{tabular} [$\uparrow$] & 
    \begin{tabular}[x]{@{}c@{}}\textbf{Entropy} \\ {[nats,$\uparrow]$} \end{tabular} \\
    \midrule
    Floating-point & $0.92\pm 0.0$ & $0.68\pm 0.01$ & $0.65\pm 0.01$ & $0.36\pm 0.14$\\
    \midrule
    Fixed-point & $0.92\pm 0.0$ & $0.68\pm 0.01$ & $0.65\pm 0.02$ & $0.38\pm 0.11$\\
    \bottomrule
  \end{tabular}}
\end{table}

Next we discuss the hardware optimization and performance comparison with respect to different hardware platforms.

\begin{table}[b]
\centering
\caption{Resource utilization for the best architectures.}
\label{utili}
\renewcommand{\arraystretch}{1.2}
\begin{tabular}{ c | c | c |c| c| c}
\toprule
 \multirow{2}{*}{\textbf{Task}}& & \textbf{LUT}   & \textbf{FF}  & \textbf{BRAM} & \textbf{DSP}\\
\cmidrule{2-6}
&  \textbf{Available} & 219k & 437k & 545 & 900  \\
\midrule
\multirow{2}{*}{\begin{tabular}[c]{@{}c@{}} Anomaly \\ $H$=16, $NL$=2, $\boldsymbol{B}$=YNYN\end{tabular}} & Used [$\downarrow$] &  207k &   218k & 149 & 758 \\ \cmidrule{2-6}
& Utilized [\%, $\downarrow$] & 94 &  49 & 13 & 84  \\
\midrule
\multirow{2}{*}{\begin{tabular}[c]{@{}c@{}} Classification \\ $H$=8, $NL$=3, $\boldsymbol{B}$=YNY \end{tabular}} & Used [$\downarrow$] &  62k &  52k & 64 & 898 \\ \cmidrule{2-6}
& Utilized [\%, $\downarrow$] & 28 & 11 & 5 & 99.8  \\
\bottomrule
\end{tabular}
\end{table}

\begin{table*}[t]
\centering
\caption{Hardware comparison between FPGA, CPU and GPU implementations.}
\label{tb:hardware_comparison}
\scalebox{0.95}{
\setlength\tabcolsep{6pt} 
\begin{tabular}{C{1.8cm}|C{1.0cm}| C{1.0cm}| C{1.0cm}| C{1.0cm}| C{1.0cm}| C{1.0cm}| C{1.0cm}| C{1.3cm}| C{1.3cm} | C{1.3cm}}
\toprule
\multirow{2}{*}{\textbf{Task}}&  {\bf Batch } & \multicolumn{3}{c|}{{\bf Latency} [ms, $\downarrow$]}& \multicolumn{3}{c|}{{\bf Power} [W, $\downarrow$]}& \multicolumn{3}{c}{{\bf Energy Consumption} [J/Sample, $\downarrow$]} \\ 
\cmidrule{3-11}
 & {\bf Size} & {FPGA}& {CPU} & {GPU}& {FPGA}& {CPU} & {GPU}& {FPGA}& {CPU} & {GPU} \\
\midrule

\multirow{2}{*}{\begin{tabular}[x]{@{}c@{}} {\scriptsize Anomaly}
\\ {\scriptsize $H$=16, $NL$=2,}\\ {\scriptsize $\boldsymbol{B}$=YNYN}\end{tabular}} 
& 50 & \textbf{41.31} & 4011 & 379.81& \multirow{2}{*}{\textbf{3.44}} & \multirow{2}{*}{15} & \multirow{2}{*}{69}& \textbf{0.005} & 2.01 & 0.53  \\
 \cmidrule{2-5}\cmidrule{9-11}
& 200 & \textbf{165.24} & 5964 & 402.76&  &  & & \textbf{0.019} & 2.98 &  0.56 \\
\midrule

\multirow{2}{*}{\begin{tabular}[x]{@{}c@{}} {\scriptsize Classification}\\ {\scriptsize $H$=8, $NL$=3,} \\ {\scriptsize $\boldsymbol{B}$=YNY}\end{tabular}} 
& 50 & \textbf{25.23} & 3690  & 245.14 & \multirow{2}{*}{\textbf{2.47}} &\multirow{2}{*}{16}  & \multirow{2}{*}{65} & \textbf{0.002} & 1.97 &  0.36 \\
 \cmidrule{2-5}\cmidrule{9-11}
& 200 & \textbf{100.92} & 4981& 256.98&  &  & & \textbf{0.008} & 2.66 & 0.38 \\

\bottomrule
\end{tabular}}
\end{table*}
\begin{table*}[t]
\centering
\caption{Optimization for anomaly detection.}
\label{tb:optimization_anomaly}
\scalebox{1.0}{
\setlength\tabcolsep{6pt} 
\begin{tabular}{C{3cm}|C{1.9cm}| C{1.0cm}| C{1.0cm}| C{1.0cm}| c| c| c}
\toprule
\multirow{2}{*}{\textbf{Mode}} &  \multirow{2}{*}{\begin{tabular}[x]{@{}c@{}}$\boldsymbol{A}:$\\ $\{H, NL, \boldsymbol{B}\}$ \end{tabular}} & \multicolumn{3}{c|}{{\bf Latency} [ms, $\downarrow$]}& \multirow{2}{*}{\textbf{Accuracy} [$\uparrow$]}  & \multirow{2}{*}{\begin{tabular}[x]{@{}c@{}}\textbf{Average}\\ \textbf{Precision}\end{tabular} [$\uparrow$]} & \multirow{2}{*}{\begin{tabular}[x]{@{}c@{}}\textbf{Area}\\ \textbf{under} \textbf{Curve}\end{tabular} [$\uparrow$]} \\ 
\cmidrule{3-5}
&  & {FPGA}& {CPU} & {GPU}&  &  &  \\
\midrule

\textit{Opt-Latency}  & $8, 1$, NN  & \textbf{6.94}  & 133.45 &  10.57&  0.93& 0.87 & 0.95   \\
\midrule
\textit{Opt-Accuracy / Precision / Area under Curve} & $16, 2$, YNYN& \textbf{165.24} & 5485  & 250.27 & \textbf{0.96}& \textbf{0.98}  & \textbf{0.99}    \\

\bottomrule
\end{tabular}}
\end{table*}
\begin{table*}[t]
\centering
\caption{Optimization for classification.}
\label{tb:optimization_classification}
\scalebox{1.0}{
\setlength\tabcolsep{6pt} 
\begin{tabular}{c|C{1.9cm}| C{0.9cm}| C{0.9cm}| C{0.9cm}| c| c| c |c}
\toprule
\multirow{2}{*}{\textbf{Mode}} &  \multirow{2}{*}{\begin{tabular}[x]{@{}c@{}}$\boldsymbol{A}:$\\ $\{H, NL, \boldsymbol{B}\}$ \end{tabular}} & \multicolumn{3}{c|}{{\bf Latency} [ms, $\downarrow$]}& \multirow{2}{*}{\textbf{Accuracy} [$\uparrow$]}  & \multirow{2}{*}{\begin{tabular}[x]{@{}c@{}}\textbf{Average}\\ \textbf{Precision}\end{tabular} [$\uparrow$]} & \multirow{2}{*}{\begin{tabular}[x]{@{}c@{}}\textbf{Average}\\ \textbf{Recall}\end{tabular} [$\uparrow$]} & \multirow{2}{*}{\textbf{Entropy} [nats, $\uparrow$]} \\ 
\cmidrule{3-5}
&  & {FPGA}& {CPU} & {GPU}&  &  &  \\
\midrule

\textit{Opt-Latency}  & $8, 1$, N   & \textbf{3.44} & 120.52 & 6.49& 0.90 & 0.62 & 0.66 &  0.15 \\
\midrule
\textit{Opt-Accuracy} & $8, 3$, NYN& \textbf{100.92} & 4799 & 193.10 & \textbf{0.93}  &  0.67 & \textbf{0.67}&  0.14  \\
\midrule
\textit{Opt-Precision} & $8, 3$, YNY &\textbf{100.92}  & 4789 & 182.59 & 0.92 & \textbf{0.69} &  0.64 & 0.30 \\
\midrule
\textit{Opt-Recall} & $8, 2$, YN& \textbf{100.91} & 3176 & 123.59 &0.91  &0.64  & \textbf{0.67} &  0.20 \\
\midrule
\textit{Opt-Entropy} & $8, 3$, YNN & \textbf{100.92} & 4795 & 191.64 & 0.89 & 0.59 & 0.64 &\textbf{0.60} \\
\bottomrule
\end{tabular}}
\end{table*}

\subsection{Performance Comparison with GPU and CPU}
We implemented the proposed design from~\secref{sec:hardware} on Xilinx ZC706, which consists of a XC7Z045 FPGA and a dual ARM Cortex-A9 processor. 1 GB DDR3 RAM is installed on the platform as the off-chip memory. The Xilinx Vivado HLS 2019.2 tool was used for synthesis. {The FPGA power is reported by the Xilinx Vivado tool.} The design frequency was 100MHz.
The reuse factors were determined through the optimization framework and set as $R_x$=16 and $R_h$=5 when $H$=16 and $R_x$=12 and $R_h$=1 when $H$=8 given the FPGA. The $R_d$ is set to $R_x$ for autoencoder and is set to 1 for classifier to achieve low latency. 

\tabref{utili} shows the resource utilization for the designs of the optimal architectures for anomaly detection and classification on the FPGA.
It can be seen that both Bayesian RNN models can fit the FPGA and almost all of the FPGA’s DSPs or LUTs were utilized with 758 and 898 DSPs used for anomaly detection or classification architectures. 
At the same time, the estimated DSP consumption for these architectures with the model presented in~\secref{sec:res_model} were 754 and 915 respectively, demonstrating fine accuracy of the resource model, which is more than 98\% accurate.

To demonstrate the advantage of our FPGA-based accelerator compared with CPU and GPU implementations,
we evaluated the best RNN models found in anomaly detection and classification tasks on the FPGA, a TITAN X Pascal {with 3,840 CUDA cores clocked at 1.4 GHz} and an Intel Xeon E5-2680 v2 CPU {with 8 CPU cores clocked at 2.4 GHz} with respect to latency, power consumption and energy consumption per sample.
{We measured the power of the CPU using a power meter. The power of GPU was reported by an Nvidia toolkit.}
PyTorch 1.8~\citep{NEURIPS2019_9015} is used for both CPU and GPU implementations. {The random number generation for CPU and GPU implementations is performed via default PyTorch calls and default pseudo-random number generators for each platform.}
To optimize the hardware performance on each platform, we use TensorRT and CuDNN $8.11$ libraries for the GPU implementation, and MKLDNN for the CPU implementation.
The number of samples was set to be $S=30$ as indicated by~\secref{sec:experiments_algorithm_optimization}.
{
As GPUs always show advantages in multi-batch workload,
we set the batch size to be $50$ and $200$ on all hardware platforms for a fair comparison. This batch size is realistic in our application, considering for example multiple patients.
}
The results are presented in~\tabref{tb:hardware_comparison}.
Compared with GPU implementation, our FPGA-based design was nearly $2{\sim}8$ times faster and consumed $20{\sim}26$ times less power.
In terms of the energy consumption, which was measured by energy per sample,
our design was nearly $106$ times more efficient than the GPU implementation.
Comparing with the CPU implementation,
our FPGA-based accelerator achieved approximately $100{\sim}400$ times higher energy efficiency. {FPGA implementations are faster and more efficient because they are unrolled on-chip with respect to our tailor-made design. 
The FPGA design processes the input with batch size 1, since requests need to be processed as soon as they arrive.}

Lastly, based on the latency model in~\secref{sec:latency_model}, 
the estimated latencies of the two architectures with 50 batches were 42.25ms and 25.77ms respectively. Hence, the analytical prediction errors were only 2.26\% and 2.13\% respectively, confirming the accuracy of the latency model.

\subsection{Optimization Framework Efficiency}
To demonstrate the effectiveness of our framework on finding optimized designs under different user-defined priorities, we evaluated the proposed framework with respect to both anomaly detection and classification on the \textit{ECG5000} dataset.

For the anomaly detection, since it is primarily a regression task, we set the optimization modes as \textit{Opt-Latency}, \textit{Opt-Accuracy}, \textit{Opt-Precision} and \textit{Opt-AUC}. {If users wish to only optimize hardware performance, they would pick the configuration with the optimal latency. 
However, if users wish to obtain a model with minimized errors, they would maximize the accuracy. 
If users wish to maximize the true positive rate and minimize the false positive rate, or in general to obtain model that has high precision on a range of thresholds, they would pick the model with the highest AUC or AP respectively.}
The results are presented in~\tabref{tb:optimization_anomaly}.
Surprisingly, we found that \textit{Opt-Accuracy}, \textit{Opt-Precision} and \textit{Opt-AUC} generated the same model with $H$=16, $NL$=2 and MCD applied in the first and third layers.
While the \textit{Opt-Latency} simply traded-off the algorithmic performance for the smallest hidden size, $NL$=1 with no MCD using $S$=1 to achieve the lowest latency.
As we can see from~\tabref{tb:optimization_anomaly}, our FPGA-based design was still $1.4{\sim}33.2$ times faster than both CPU and GPU implementations depending on different model architectures.

For the classification task, there can be up to five optimization modes, namely \textit{Opt-Latency}, \textit{Opt-Accuracy}, \textit{Opt-Precision}, \textit{Opt-Recall} and \textit{Opt-Entropy}. In addition to the modes presented in the previous paragraph, if the users wish to minimize false negatives, e.g. diagnosing a normal condition in an anomalous ECG, they would pick the model with the highest recall. 
If the model has to support high uncertainty containing outlier ECG signal values, the user could pick the model with the highest entropy.
Different optimization modes generated different model architectures as shown in \tabref{tb:optimization_classification}. 
The highest accuracy we can achieve was $93$\%. 
Similarly, with the help of our framework, we achieved $0.68$ AP, $0.67$ AR and $0.60$ nats entropy
under different optimization modes with the speedup ranging from $1.2$ to $1.9$ compared to GPU implementations.
Again, the \textit{Opt-Latency} traded-off the algorithmic performance for the smallest hidden size, non-Bayesian architecture with $NL$=1 and $S$=1 to target the improvement in the hardware performance.
Note that, although the models with $NL$=3 or $NL$=2 LSTM layers in~\tabref{tb:optimization_classification} achieve similar latency as discussed in~\secref{sec:latency_model} due to pipelining, their resource and energy consumption are different.

\section{Conclusion}\label{sec:conclusion}
This work proposes a novel high-performance FPGA-based design to accelerate Bayesian LSTM-based recurrent neural networks inferred through Monte Carlo Dropout. The presented design is sufficiently versatile to support a variety of network models with respect to different safety-critical tasks concerning real-time performance on analyzing electrocardiograms. In comparison to the GPU implementation, our FPGA-based design can achieve up to 10 times speedup with nearly 106 times higher energy efficiency. At the same time, this is the first work that is focused on accelerating Bayesian recurrent neural networks on an FPGA. Moreover, this work presents an end-to-end framework to automatically trade-off both algorithmic and hardware performance, given algorithmic requirements and hardware constraints.
In future work we aim to explore co-design of custom recurrent cells
and reconfigurable hardware accelerators, to obtain the most
optimized configurations and hardware implementations. {Additionally, we are interested in supporting a wide variety of dropout rates in hardware.}

\section*{Acknowledgment}

This work was supported in part by the United Kingdom EPSRC under Grant EP/L016796/1, Grant EP/N031768/1, Grant EP/P010040/1, Grant EP/V028251/1 and Grant EP/S030069/1
and in part by the funds from Corerain, Maxeler, Intel, Xilinx and State key lab of Space-Ground Integrated Information Technology (SGIIT). Martin Ferianc was sponsored through a scholarship from the Institute of Communications and Connected Systems at UCL. We also thank the reviewers for insightful comments and suggestions.

\printbibliography

\end{document}